\documentclass[runningheads]{llncs}

\usepackage{eccv}

\usepackage{eccvabbrv}

\usepackage{graphicx}
\usepackage{booktabs}
\usepackage{algorithm}
\usepackage{algorithmic}
\usepackage{amsmath}

\usepackage[accsupp]{axessibility}  % Improves PDF readability for those with disabilities.

\usepackage{hyperref}

\let\svthefootnote\thefootnote
\newcommand\freefootnote[1]{%
  \let\thefootnote\relax%
  \footnotetext{#1}%
  \let\thefootnote\svthefootnote%
}

\usepackage{orcidlink}

\usepackage{marvosym}
\usepackage{booktabs}
\usepackage{multirow}

\begin{document}

% ---------------------------------------------------------------
% TODO REVIEW: Replace with your title
\title{Post-training Quantization with Progressive Calibration and Activation Relaxing for Text-to-Image Diffusion Models} 

% TODO REVIEW: If the paper title is too long for the running head, you can set
% an abbreviated paper title here. If not, comment out.
% \titlerunning{Abbreviated paper title}
\titlerunning{PCR}

% TODO FINAL: Replace with your author list. 
% Include the authors' OCRID for the camera-ready version, if at all possible.
\author{Siao Tang\inst{1,3} \and
Xin Wang\inst{2}\orcidlink{0000-0002-0351-2939} \textsuperscript{\Letter} \and
Hong Chen\inst{2} \and
Chaoyu Guan\inst{2,3} \and
Zewen Wu\inst{2} \and
Yansong Tang\inst{1} \and
Wenwu Zhu\inst{2}\orcidlink{0000-0003-2236-9290} \textsuperscript{\Letter} 
}

% TODO FINAL: Replace with an abbreviated list of authors.
\authorrunning{S. Tang et al.}
% First names are abbreviated in the running head.
% If there are more than two authors, 'et al.' is used.
% TODO FINAL: Replace with your institution list.
\institute{TBSI, Shenzhen International Graduate School, Tsinghua University \and 
Department of Computer Science and Technology, BNRIST, Tsinghua University
\and
Tsingmao  Intelligence \\ 
\email{\{tsa22, h-chen20, guancy19, wuzw21\}@mails.tsinghua.edu.cn} \\ 
\email{tang.yansong@sz.tsinghua.edu.cn}  \\
\email{\{xin\_wang, wwzhu\}@tsinghua.edu.cn}  
% \email{lncs@springer.com}\\
% \url{http://www.springer.com/gp/computer-science/lncs} \and
% ABC Institute, Rupert-Karls-University Heidelberg, Heidelberg, Germany\\
% \email{\{abc,lncs\}@uni-heidelberg.de}
}

\freefootnote{{\Letter} denotes corresponding authors.}

\maketitle

\begin{abstract}
High computational overhead is a troublesome problem for diffusion models. Recent studies have leveraged post-training quantization (PTQ) to compress diffusion models. However, most of them only focus on unconditional models, leaving the quantization of widely-used pretrained text-to-image models, e.g., Stable Diffusion, largely unexplored. In this paper, we propose a novel post-training quantization method PCR (Progressive Calibration and Relaxing) for text-to-image diffusion models, which consists of a progressive calibration strategy that considers the accumulated quantization error across timesteps, and an activation relaxing strategy that improves the performance with negligible cost. Additionally, we demonstrate the previous metrics for text-to-image diffusion model quantization are not accurate due to the distribution gap. To tackle the problem, we propose a novel QDiffBench benchmark, which utilizes data in the same domain for more accurate evaluation. Besides, QDiffBench also considers the generalization performance of the quantized model outside the calibration dataset. Extensive experiments on Stable Diffusion and Stable Diffusion XL demonstrate the superiority of our method and benchmark. Moreover, we are the first to achieve quantization for Stable Diffusion XL while maintaining the performance.
\end{abstract}

\section{Introduction}
\label{sec:intro}

Diffusion models have demonstrated remarkable ability in various generative tasks~\cite{ho2020denoising,song2020denoising,rombach2022high,chen2023disenbooth,ruiz2023dreambooth,song2020score,10579040}, especially in text-to-image generation~\cite{rombach2022high,chen2023disenbooth,ruiz2023dreambooth}. However, one critical limitation of diffusion models is the high computational cost, which mainly results from the following two aspects. On the one hand, diffusion models typically require multiple denoising steps to generate images, which is time-consuming. On the other hand, the large network architecture of diffusion models usually requires a great deal of time and memory, especially for foundation models pretrained on large-scale datasets, e.g., Stable Diffusion~\cite{rombach2022high} and Stable Diffusion XL~\cite{podell2023sdxl}. In this paper, we focus on the latter problem, i.e., compressing the diffusion models to reduce the computational cost.

Recent studies have explored quantization methods to compress diffusion models~\cite{shang2023post,li2023q, he2023ptqd,wang2023towards,so2023temporal}, which mainly adopt Post-training Quantization (PTQ)~\cite{nagel2020up,li2021brecq}. PTQ has no need for retraining or finetuning the network, making it more resource-friendly than Quantization-aware Training (QAT)~\cite{tailor2020degree,nagel2022overcoming}. Nevertheless, most of these previous works focus on the unconditional diffusion models, leaving the quantization of the widely used large pretrained text-to-image models, e.g., Stable Diffusion, largely unexplored. Only Q-diffusion~\cite{li2023q} presents several quantitative quantization results for the text-to-image diffusion model. However, the results are limited because of the adopted inaccurate metrics without considering the data distribution-gap problem.

In this paper, we investigate the largely unexplored text-to-image diffusion model quantization problem, from both method and benchmark perspectives.

On the one hand, about the method, all previous works use the full-precision model to obtain the calibration data for each timestep, and thus ignore the accumulated quantization error of previous denoising steps, which may result in inaccurate quantization. 
% Additionally, they neglect the effects of different denoising steps on image fidelity or text-to-image matching, which may largely influence the quantization performance.
Additionally, they neglect the sensitivity of image fidelity or text-image matching to different denoising steps, which may largely influence the quantization performance.
On the other hand, about the metrics, previous works use a prompt subset of the COCO~\cite{lin2014microsoft} dataset to perform the calibration and use the FID metric calculated with COCO images to evaluate the model. However, this metric leads to inaccurate evaluation because there exists a distribution gap between the generated images by the text-to-image diffusion model (pretrained on large-scale datasets) and the COCO images. Additionally, previous metrics only test on the COCO prompts, ignoring the prompt-generalization ability to unseen prompts outside the calibration dataset, which is important to the pretrained text-to-image diffusion models.

To tackle the problems, we propose a novel quantization method PCR (Progressive Calibration and Relaxing) and the QDiffBench benchmark for quantizing text-to-image diffusion models. Specifically, the proposed PCR method consists of a progressive calibration strategy and an activation relaxing strategy. The progressive calibration progressively quantizes each step with all the previous steps quantized, which can be aware of the accumulated quantization error across steps. The activation relaxing strategy relaxes a minor part of critical sensitive timesteps using higher bitwidth to improve image fidelity or text-image matching, with negligible additional cost.

As for benchmark, QDiffBench consists of an accurate FID calculation strategy and a prompt-generalization evaluation strategy. To deal with the data distribution-gap problem, QDiffBench calculates the FID~\cite{heusel2017gans} score between the images generated by the full-precision model and the quantized model. This strategy uses the data in the same domain for evaluation, making it more accurate in evaluating the 
% fidelity  
degradation caused by quantization. Moreover, QDiffBench evaluates the generalization performance for the quantized text-to-image diffusion model on the prompts which have quite different styles from the calibration prompts. We emphasize the significance of QDiffBench, as the quantization for text-to-image diffusion models could be blocked without an effective benchmark.

% High-resolution text-to-image generation is important in real-world applications. Besides Stable Diffusion, we also conduct experiments on Stable Diffusion XL which is stronger and 4 times larger. Experimental results show that our method causes no CLIP score loss for Stable Diffusion, and has comparable performance to the full-precision model for Stable Diffusion XL.

To summarize, our contributions are listed as follows:

\begin{enumerate}
    \item We propose a novel PTQ quantization method \textbf{PCR} for text-to-image diffusion models, which consists of a progressive calibration strategy and an activation relaxing strategy.
   
    \item We propose a comprehensive and effective benchmark \textbf{QDiffBench}, and to our knowledge, it is the first effective benchmark to evaluate quantized text-to-image diffusion models.
    % for the quantization of text-to-image diffusion models. 

    % It adopts a more accurate FID scheme to evaluate the image fidelity, and emphasizes the prompt-generalization ability of the quantized model.

    \item Extensive experiments on foundation diffusion models (i.e., Stable Diffusion and Stable Diffusion XL) demonstrate the superiority of our proposed method and benchmark. 

    \item We are the first to achieve quantization on Stable Diffusion XL (3.5B parameters) which is one of the biggest diffusion models.
    % and generate high-quality quantized images 
    % with higher resolution than 512 × 512.
\end{enumerate}

\section{Related Work}
\label{sec: related work}

\subsection{Model Quantization}

Model quantization~\cite{nagel2021white} is one of the most significant techniques of model compression. Quantization methods compress neural networks by compressing the model weights and the activations to lower bits, e.g., converting 32/16-bit float weights/activations into formats with INT4 or INT8 bit-width. The quantization process can be formulated as:

\begin{align}
w_q= \operatorname{clip}\left(\operatorname{round}\left(\frac{w}{s}\right)+z, q_{\min }, q_{\max }\right), \nonumber
\end{align}

\noindent where $s$ is the scaling factor and $z$ is the zero-point. 

% The corresponding de-quantization process is:
% \begin{align}
% \hat{w}= s \cdot \left(w_q-z\right). \nonumber
% \end{align}

\textbf{Post-training Quantization.} Quantization methods are usually divided into two classes, i.e., post-training quantization (PTQ)~\cite{nagel2020up,li2021brecq}  and quantization-aware training (QAT)~\cite{tailor2020degree,nagel2022overcoming}. QAT methods involve training/finetuning, usually requiring massive training data and costing many computation resources. In contrast, PTQ methods only require a small subset of data (e.g., 128 images) for calibration, and have no need for training, making them more resource-friendly and faster than QAT methods.
% PTQ methods have been widely studied for LLMs~\cite{frantar2022gptq,xiao2023smoothquant,yao2022zeroquant,bai2022towards}.
% Recent studies have explored quantization methods
% for diffusion models~\cite{shang2023post,li2023q, he2023ptqd,wang2023towards,so2023temporal}, mostly focusing on PTQ.

\textbf{Mixed-precision Quantization.} Mixed-precision quantization~\cite{wu2018mixed,dong2019hawq} is a strategy that allocates different bitwidth to different layers/blocks, which is considered a trend~\cite{wang2019haq,koryakovskiy2023one} because it can further improve the model performance. Mixed-precision quantization can be applied in practical scenes, as more and more hardwares support it, such as Qualcomm Snapdragon 8 Gen 2~\cite{qualcomm}, NVIDIA A100~\cite{choquette2021nvidia}, and RTX 4090~\cite{rtx4090}. Our activation relaxation strategy can also be regarded as a mixed-precision quantization in the dimension of timestep.

\subsection{Quantization for Diffusion Model}

Quantization methods have been used to compress and accelerate diffusion models~\cite{shang2023post,li2023q, he2023ptqd,wang2023towards,so2023temporal}.
For example, Shang et al. propose PTQ4DM~\cite{shang2023post} which collects calibration data in the denoising sampling process rather than simulating the training process. They also propose that the timesteps of calibration data should be generated from a skew-normal distribution. Q-diffusion~\cite{li2023q} proposes to collect calibration data at certain timestep intervals and split quantization for shortcut layers whose input is a concatenated skip-connection feature. However, all the existing quantization methods ignore the accumulated error across sampling steps. Additionally, none of them tailor the design for the setting of text-to-image generation.

\section{Method}

In this section, we present our method \textbf{PCR}, which consists of a progressive calibration strategy that considers the accumulated quantization error across timesteps, and an activation relaxing strategy which can effectively improve performance with negligible costs. We also present the pseudo-code of PCR in \cref{algorithm: pcr} (in the Appendix).

\subsection{Preliminaries for Diffusion Models}

Diffusion models define a forward process that adds the noise to the real data $\mathbf{x}_0$:

{
\begin{align}
\mathbf{x}_{t} = \alpha_t \mathbf{x}_{0} +  \sigma_t \boldsymbol{\epsilon}, \ \boldsymbol{\epsilon} \sim \mathcal{N}(\mathbf{0}, \mathbf{I}), \nonumber
\end{align}
}

\noindent where $\alpha_t$ and $\sigma_t$ are functions of $t$ and the signal-to-noise-ratio $\alpha_t / \sigma_t$ is strictly decreasing. Diffusion models train a neural network ${\boldsymbol{\epsilon}_{\theta}}$ to predict the noise of a noisy variable $\mathbf{x}_{t}$, whose training objective is:

\begin{align}
\mathbf{E}_{\mathbf{x}_{0}, \boldsymbol{\epsilon}, t} 
[\left\|\boldsymbol{\epsilon}_{\theta}\left(\alpha_{t} \mathbf{x}_{0}+\sigma_{t} \boldsymbol{\epsilon},t \right)-\boldsymbol{\epsilon}\right\|_{2}^{2}]  , \nonumber
\end{align}

In the reverse process, diffusion models generate samples by gradually denoising from a Gaussian noise $\mathbf{x}_T \sim \mathcal{N}(\mathbf{0}, \mathbf{I})$ to $\mathbf{x}_0$, along certain trajectories determined by sampling strategies~\cite{ho2020denoising, liupseudo, lu2022dpm}.

\subsection{Time-Accumulated Error Aware Progressive Calibration}
\label{sec: progressive calibration}

Substantial studies~\cite{shang2023post,li2023q,so2023temporal, wang2023towards} point out that the distributions of activation obviously vary over different time steps, which is a challenge for quantizing diffusion models. To tackle this problem, some studies~\cite{so2023temporal, wang2023towards} adopt time-aware quantization, i.e., separately quantizing the activations for each timestep, which has been shown to be effective. We also adopt the time-aware strategy.

\begin{figure}
    \centering
    \includegraphics[width=0.63\linewidth]{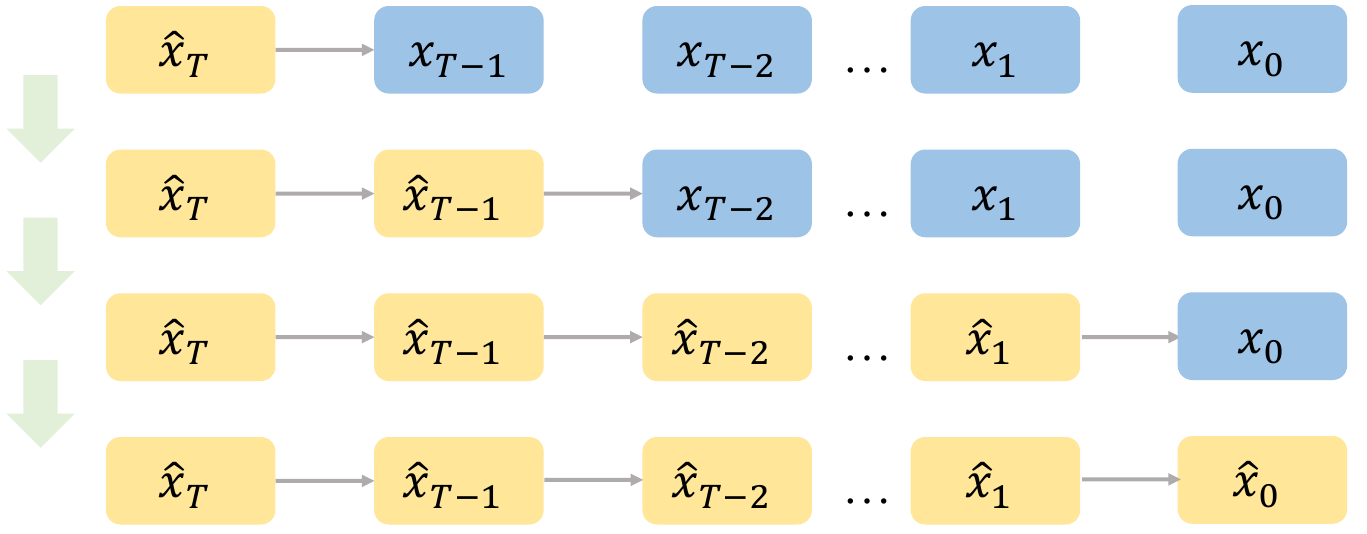}
    \caption{Progressive calibration for activations. The yellow bar means the activation has been quantized. We quantize the activations at step $t$ with all the previous steps quantized, which considers the accumulated quantization error across timesteps.}
    \label{fig:prog}
\end{figure}

\begin{figure}
    \centering
    \includegraphics[width=0.8\linewidth]{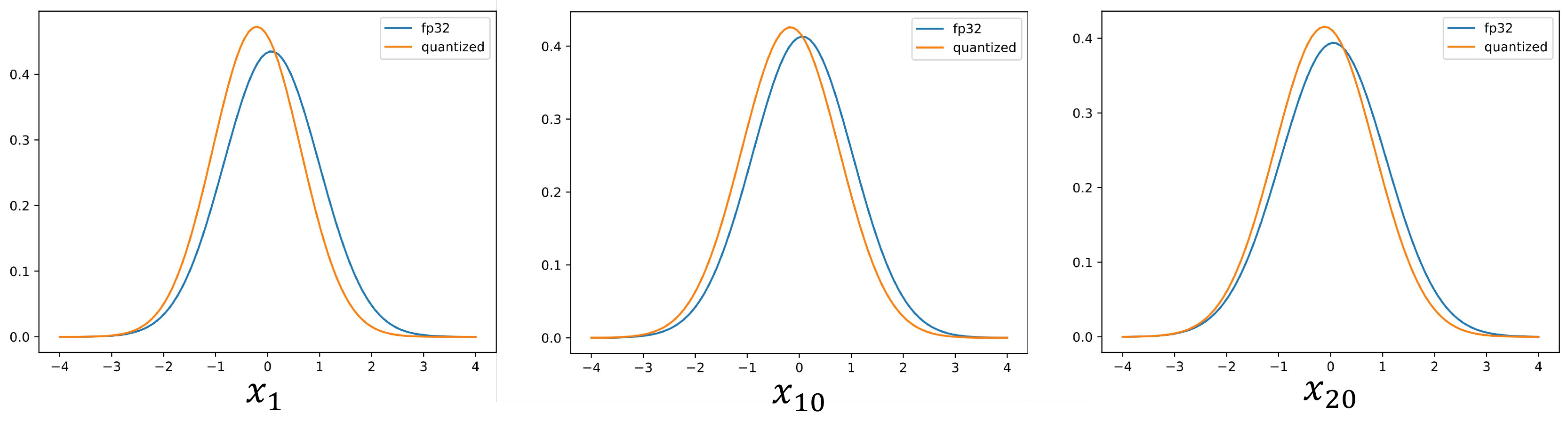}
    \caption{The distribution shift caused by the accumulated quantization error. We gather the statistics of sub feature maps at timestep 1, 10, and 20. }
    \label{fig: density}
\end{figure}

Previous works use the full-precision model to obtain the calibration data for each timestep during the sampling process.
However, they are not aware of the accumulated quantization error across timesteps in the iterative denoising process. The quantization errors in previous sampling steps will
shift the distribution of the activations in subsequent steps, as shown in \cref{fig: density}, which makes calibrating with the full-precision model sub-optimal.
% the inconsistency between calibration and inference. 
To tackle this problem, we propose a progressive calibration strategy for activations, which quantizes the activations at step $t$ with all the previous steps $t+1,..,T$ quantized, illustrated in \cref{fig:prog}.

We further mathematically prove the necessity of considering the accumulated quantization errors, through \cref{theorem:1} and its \cref{corollary:1}. The detailed proof of \cref{theorem:1} is shown in \cref{appdedix: proof}(Appendix).

\noindent \textbf{Notations.} Let $\boldsymbol{\epsilon}_{\theta}$ be the denoising net and $\mathbf{x}_t$ be the intermediate variable at timestep $t$. Let the mark $\hat{..}$ denote the quantized term, and $\hat{\boldsymbol{\epsilon}}_{\theta}\left(\hat{\mathbf{x}}_{t}, t\right)=\boldsymbol{\epsilon}_{\theta}\left(\hat{\mathbf{x}}_{t}, t\right)+\Delta_{t}$, where $\boldsymbol{\epsilon}_{\theta}\left(\hat{\mathbf{x}}_{t}, t\right)$ is the predicted noise with quantized input but full-precision network, and $\hat{\boldsymbol{\epsilon}}_{\theta}\left(\hat{\mathbf{x}}_{t},t\right)$ is the predicted noise with both quantized input and network. 
% Without loss of generality, we use the DDPM sampler there, i.e.,

% \begin{align}
% \mathbf{x}_{t-1}=\frac{1}{\sqrt{\alpha_{t}}}\left(\mathbf{x}_{t}-\frac{1-\alpha_{t}}{\sqrt{1-\bar{\alpha}_{t}}} \boldsymbol{\epsilon}_{\theta}\left(\mathbf{x}_{t}, t\right)\right)+\sigma_{t} \mathbf{z} . \nonumber
% \end{align}

\begin{theorem}
Through the multi-step denoising process, the upper bound of the ultimate quantization error $\delta = \|\mathbf{x}_0 - \mathbf{\hat{x}}_0\|$ can be approximated as the linear combination of $\{\|\Delta_t\|\}, \ t=1,..,T.$
\label{theorem:1}
\end{theorem}

\begin{corollary}
To minimize the upper bound of the final quantization error $\delta = \|\mathbf{x}_0 - \hat{\mathbf{x}}_0\|$, we should minimize the quantization error $\|\Delta_t\|$ at each timestep $t, \ t=1,..,T.$
\label{corollary:1}
\end{corollary}

\noindent Note that, $\|\Delta_{t}\|=\|\hat{\boldsymbol{\epsilon}}_{\theta}\left(\hat{\mathbf{x}}_{t}, t\right)-\boldsymbol{\epsilon}_{\theta}\left(\hat{\mathbf{x}}_{t}, t\right)\|$ is related to the quantized input $\hat{\mathbf{x}}_t$ rather than the full-precision input $\mathbf{x}_t$. Hence, to minimize $\|\Delta_{t}\|$, we should use the data generated with all its previous steps quantized as the calibration data.

% \noindent where $x_T$ equals to $\hat{x}_T$ , since it is the first step of the sampling process. 
% The equation $(a)$ holds since $x_T$ equals to $\hat{x}_T$ at the first step. Similarly, 
% holds because 
% $x_T$ equals to $\hat{x}_T$ with no previous steps.

\subsection{Time-wise Activation Relaxing}

\begin{figure}
    \centering
    \includegraphics[width=0.75\linewidth]{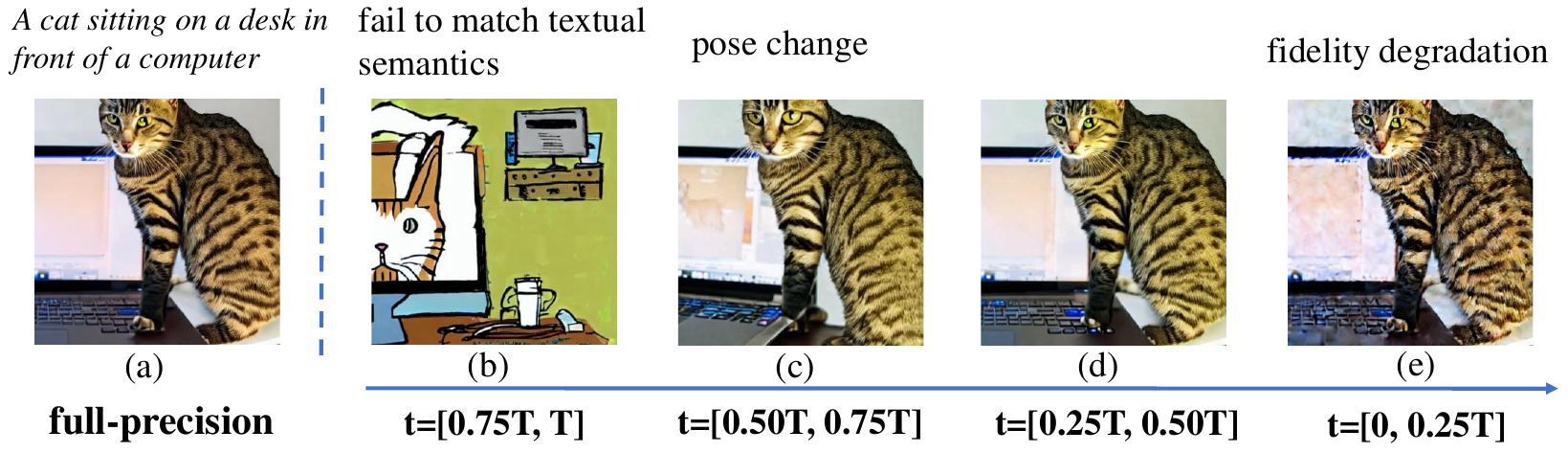}
    \caption{Add perturbations to the predicted noise at timesteps in the interval t=[a,b]. (b) shows perturbations at steps near $x_T$ make the image fail to match the prompt. (c) and (d) shows perturbations at the middle steps cause minor semantic changes such as color and computer pose. (e) shows the fidelity degradation brought by perturbations at steps near $x_0$. }
    \label{fig:visual_add_noise}
\end{figure}

Firstly, we clarify the definitions of ``text-image matching'' and ``image fidelity''. Text-image matching means whether the semantics of the image matches its prompt, which is the high-level information. In contrast, image fidelity denotes whether the image has some low-level losses such as distortions and noises.

We observe the image fidelity and text-image matching of text-to-image diffusion models are sensitive to activation quantization. To identify this sensitivity with respect to timesteps, we manually add a small Gaussian perturbation to the predicted noise at certain timesteps, simulating quantization errors. \cref{fig:visual_add_noise} shows the visual results, from which we can draw the conclusion that: \textit{Image fidelity is sensitive to the timesteps near $x_0$, while text-image matching is sensitive to the steps near $x_T$.} We refer to this phenomenon as \textit{sensitivity discrepancy}. 

% We observe the image fidelity and text-image matching of text-to-image diffusion models are sensitive to activation quantization. To identify this sensitivity with respect to timesteps, we manually add a small perturbation based on quantization error to the predicted noise at certain timesteps, simulating quantization errors. \cref{fig:visual_add_noise} shows the visual results, from which we can draw the conclusion that: \textit{Image fidelity is sensitive to the timesteps near $x_0$, while text-image matching is sensitive to the steps near $x_T$.} We refer to this phenomenon as \textit{sensitivity discrepancy}. 

\begin{figure}
    \centering
    \includegraphics[width=0.5\linewidth]{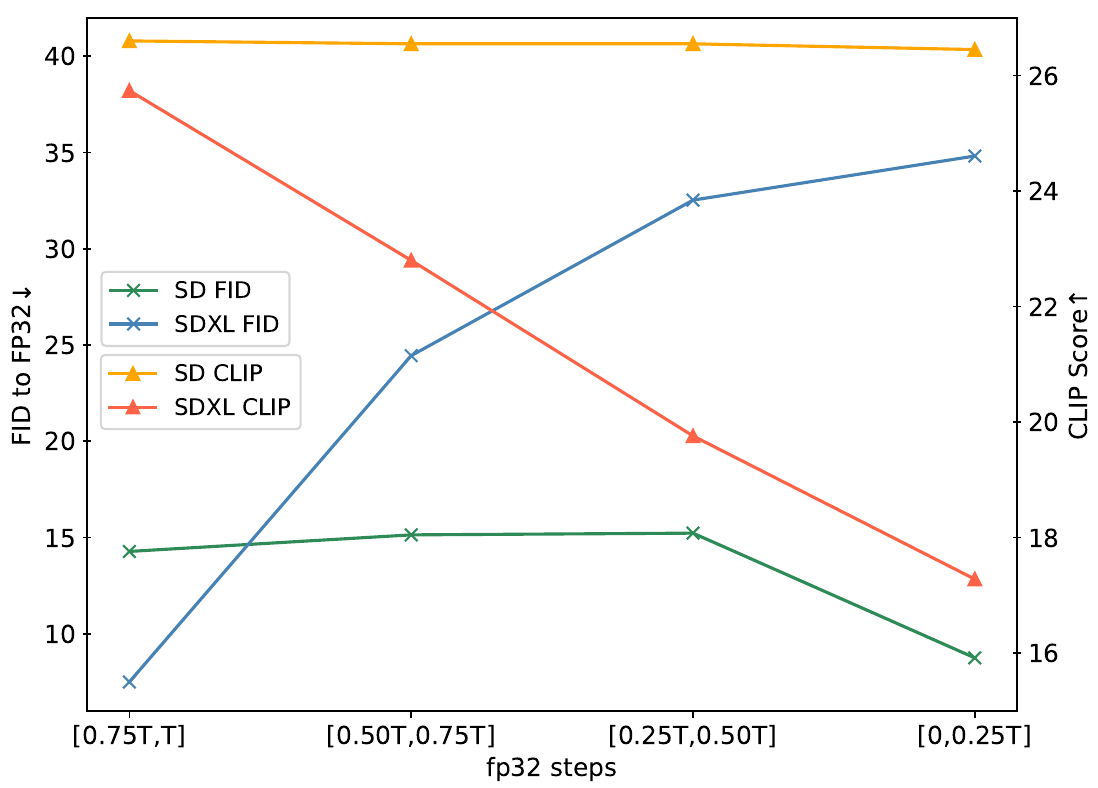}
    \caption{Quantitative results of setting certain steps to full-precision. CLIP score measures the quality of text-image matching. FID to FP32 (presented in \cref{benckmark: fid to fp32}) measures image fidelity when text-image matching is good, but it can be quite poor when there is a bad text-image matching.  } 
    \label{fig:sensitivity}
\end{figure}

To further identify actual patterns of sensitivity discrepancy in practical quantization scenes, we first quantize a model to low-bit and then set certain timesteps to full-precision. According to the results in~\cref{fig:sensitivity}, we observe that \textit{Stable Diffusion is sensitive to image fidelity degradation (controlled by steps near $x_0$) rather than text-image matching degradation  (controlled by steps near $x_T$), while Stable Diffusion XL is the opposite. }

However, previous works neglect this sensitivity discrepancy, and allocate the same bitwidth to all timesteps, resulting in a sub-optimal solution. To tackle this issue, we propose to relax a minor part of timesteps using higher bitwidth to improve image fidelity or text-image matching. Let ${b_t}, t =0,.., T$ denote the bit widths for activation quantization at step $t$. For the quantized models which are sensitive to image fidelity degradation (e.g., Stable Diffusion), we relax the $m$ steps near $x_0$ by setting ${b_i}, i =0,.., m-1$ to a higher bitwidth such as 10-bit. For the models that are sensitive to text-image matching (e.g., Stable Diffusion XL), we set ${b_i}, i =T-m+1,.., T$ to a higher bit. 
The relaxation proportion $\tau=m/T$ conforms to the law of diminishing marginal utility (\cref{ablation study}), so we propose to set $\tau$ no more than 0.20 for the trade-off between performance and extra computation. To identify the sensitivity of a given model to image fidelity or text-matching, users can perform simple experiments as \cref{fig:sensitivity}, which only takes several minutes since we adopt PTQ.

\begin{figure}
    \centering
    \includegraphics[width=0.79\linewidth]{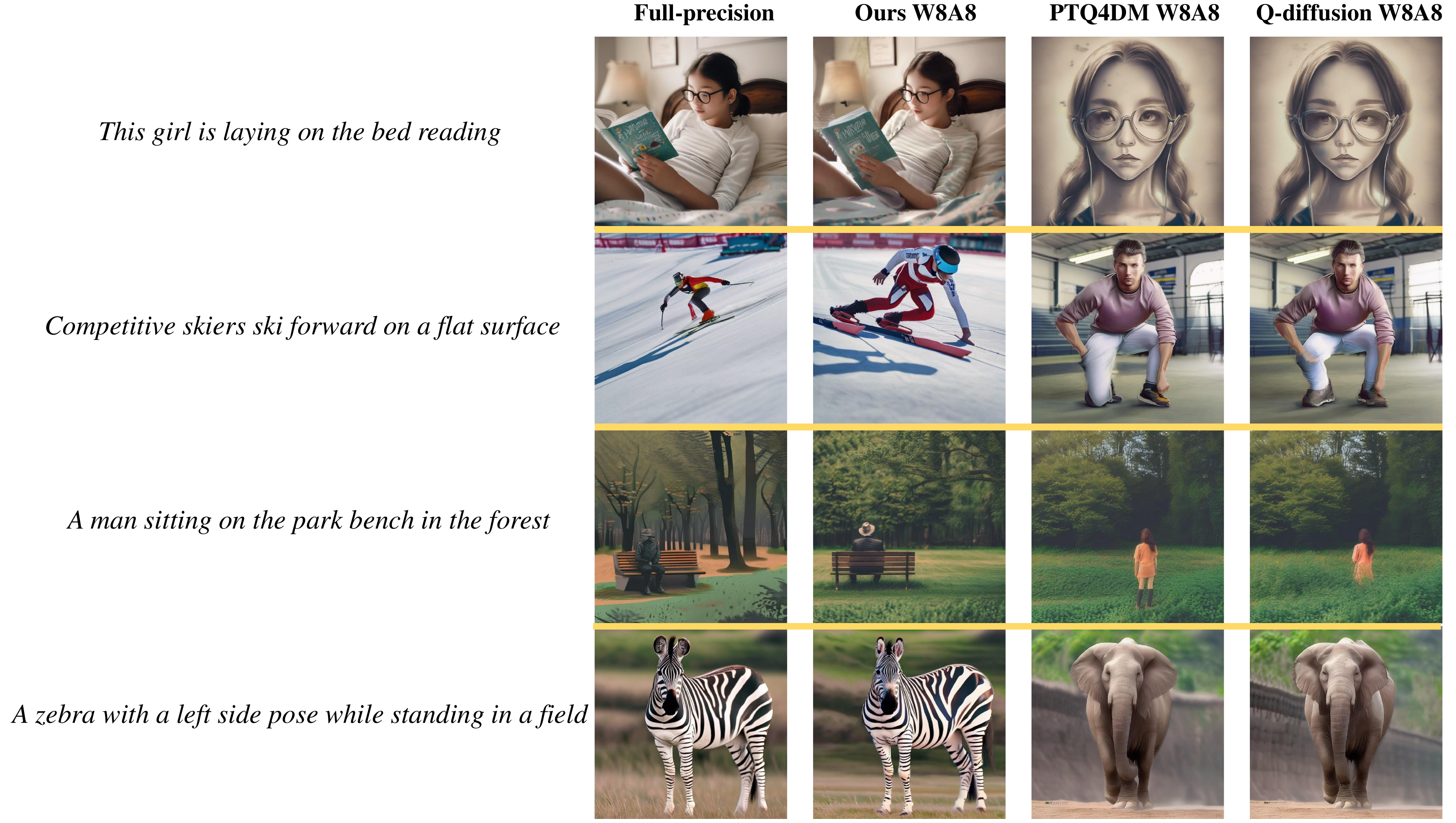}
    \caption{Stable Diffusion XL 768x768 generation results. PTQ4DM and Q-diffusion fail to match the textual semantics, while our method with the activation relaxing can generate high-quality images matching the prompt.}
    \label{fig:sdxl comparison}
\end{figure}

\cref{fig:sdxl comparison} shows previous methods lose the ability to match the textual semantics in Stable Diffusion XL, while our method with relaxation maintains it well.

Although the relaxation remarkably improves the performance, it only requires negligible costs, since we only relax a small proportion of timesteps that are the most critical. For instance, in 8-bit quantization, relaxing 5\% steps to 10 bits while keeping others the 8-bit actually results in the average 8.1 bit.

% In our experiments, we obtain the following observation:

% \textit{Stable Diffusion is sensitive to image fidelity degradation brought by quantization, while Stable Diffusion XL is sensitive to text-image matching.}

% With the observation, we relax the steps near $x_0$ for Stable Diffusion. In contrast, Stable Diffusion XL is more sensitive to text-image matching, which might be the result of using massive cross-attention transformer blocks. Most images generated by the quantized Stable Diffusion XL with previous methods can not properly match the textual semantics, shown in \cref{fig:sdxl comparison}. To tackle this problem, we relax the timesteps near $x_T$ and can generate high-quality images matching the text.

% \begin{figure}
%     \centering
%     \includegraphics[width=0.8\linewidth]{sec/imgs/SDXL-comparison-1.pdf}
%     \caption{Stable Diffusion XL 768x768 generation results. PTQ4DM and Q-diffusion fail to match the textual semantics, while our method with the activation relaxing can generate high-quality images matching the prompt.}
%     \label{fig:sdxl comparison}
% \end{figure}

\section{QDiffBench}

Previous works rarely test their methods on text-to-image generation tasks. Only Q-diffusion~\cite{li2023q} reports few quantitative results of Stable Diffusion. However, we observe that there exist some problems in the way of calculating FID used in Q-diffusion, which is caused by distribution-gap.
Moreover, they ignore the prompt-generalization ability of the foundation text-to-image model. In this section, we discuss these issues of previous metrics and propose a novel benchmark QDiffBench, which consists of a more accurate FID strategy and a prompt-generalization evaluation strategy, to accurately and comprehensively evaluate the quantization of text-to-image diffusion models.

\subsection{FID to Full Precision Model}
\label{benckmark: fid to fp32}

FID ((Frechet Inception Distance)~\cite{heusel2017gans} is an important metric to evaluate the quality of generative models, which calculates the distance between the generated images and the real images, where a lower score indicates better performance. FID is also usually used to evaluate text-to-image diffusion models~\cite{rombach2022high}. For example, Rombach et al.~\cite{rombach2022high} evaluate latent text-to-image diffusion models on the COCO validation dataset. They first generate images conditioned by COCO prompts, and calculate the FID score between the generated images and real COCO images.

\begin{figure*}
    \centering
    \includegraphics[width=1\linewidth]{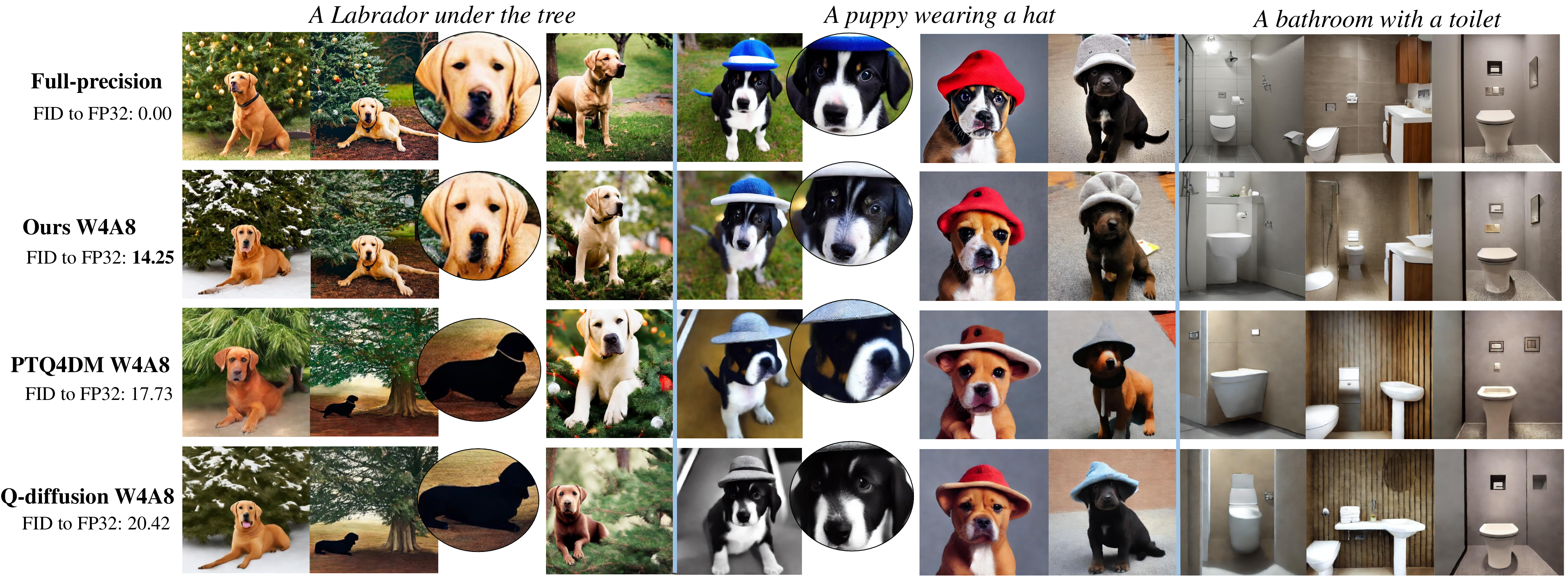}
    \caption{Comparison between the quantized models and the full-precision model. Use Stable diffusion 512 x 512. Relaxation proportion $\tau=0.20$.}
    \label{fig:sd hand prompts}
\end{figure*}

Previous studies also use the same way to evaluate quantized text-to-image diffusion models~\cite{li2023q}, i.e., generating
images with a quantized model and calculating FID between the generated and the real ones. We name this strategy ``FID to COCO''. However, we demonstrate this strategy is inaccurate in the context of quantization. 
For instance, as shown in \cref{table: coco-results}, the quantized models usually have similar or even lower ``FID to COCO'' scores than the full-precision model. However, the generated images by the quantized models are generally worse than the precision ones, taking \cref{fig:sd hand prompts} as an example. In addition, \cref{table: coco-results} also shows the W4A8 models have better ``FID to COCO'' scores than the W8A8 peers, which is also inconsistent with the visualized results, as shown in the Appendix (\cref{fig:appendix sd on coco}).
These phenomena prove the inaccuracy of ``FID to COCO''.
The reason is that the distribution of the generated images by text-to-image diffusion models obviously differs from the COCO images.
% mainly because the models are pretrained on large-scale datasets such as LAION~\cite{schuhmann2022laion}.
% One apparent manifestation of this distribution gap is the style difference.
The manifestations of this distribution gap include style differences, fake details in generated images, etc.
We detailedly analyze the reasons and manifestations of distribution-gap in the Appendix (\cref{sec: analysis about distribution gap}).
This gap makes the distance to the COCO images inaccurate to measure the quality of the images generated by quantized models. Therefore, the worse quantized models can even have better ``FID to COCO'' scores.

\begin{figure}
    \centering
    \includegraphics[width=0.6\linewidth]{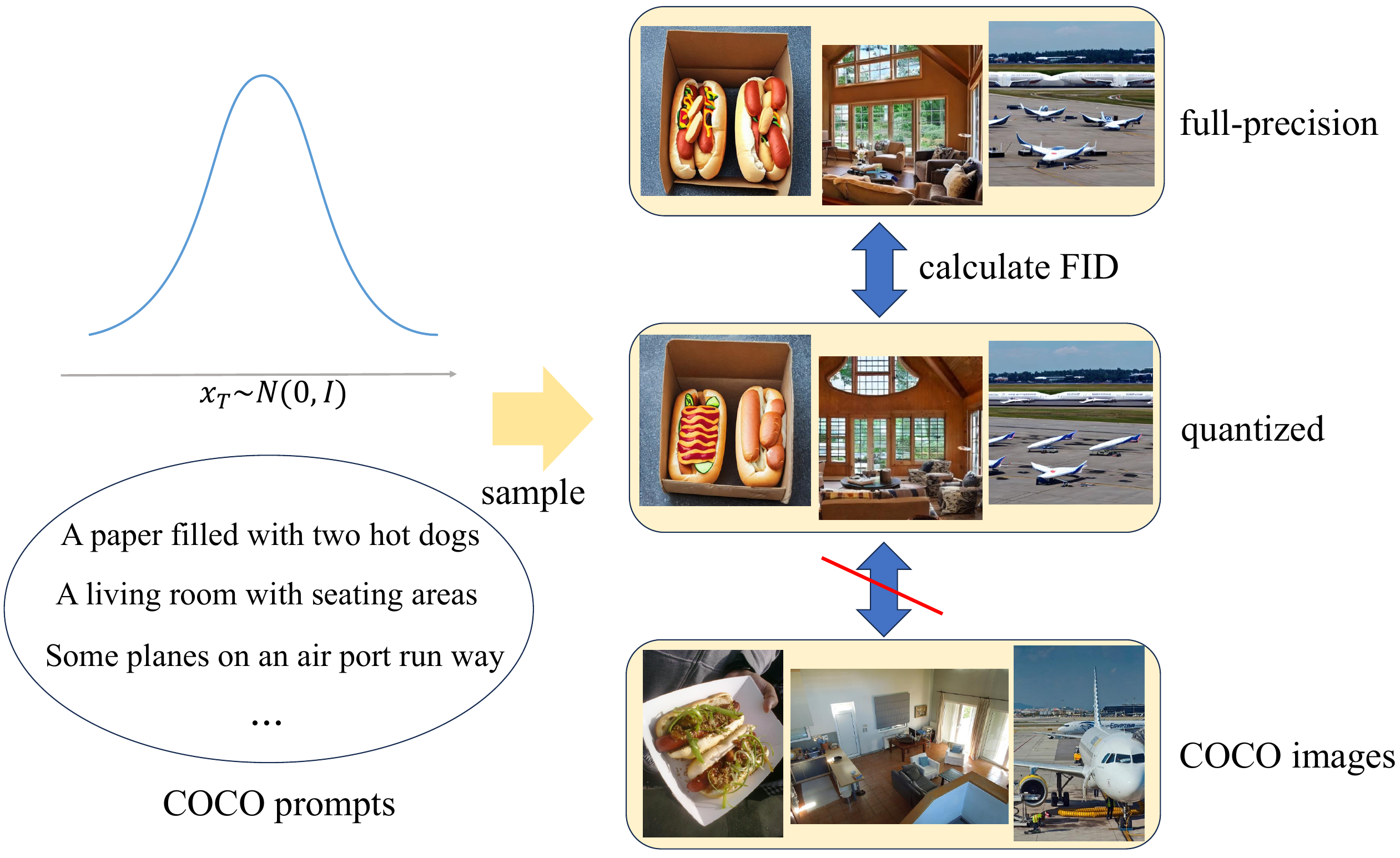}
    \caption{We calculate the FID to the images generated by the full-precision model rather than the COCO images. COCO images are photorealism, while Stable Diffusion can't achieve this perfectly. Additionally, there are obvious fake details in the generated plane images.}
    \label{fig: fid metric}
\end{figure}

To tackle this problem, we propose to directly compare the images generated by the quantized model with the images generated by the full precision model, i.e., calculating the FID score between them, as shown in \cref{fig: fid metric}. We name this strategy ``FID to FP32''. This strategy straightforwardly measures the
% fidelity
degradation caused by the quantization algorithm. It compares the generated images to the data in the same domain, making it more accurate. 
FID to FP32 measures image fidelity when text-image matching is good, but it can be quite poor with a bad text-image matching, 

\cref{table: coco-results} also shows our ``FID to FP32' metric is basically consistent with the CLIP score. \cref{fig:sd hand prompts} and the samples in the Appendix also visually demonstrate our metric is consistent with human cognition.

\subsection{Prompt-Generalization Ability}
\label{benckmark:ood}

\begin{figure*}
    \centering
    \includegraphics[width=1.0\linewidth]{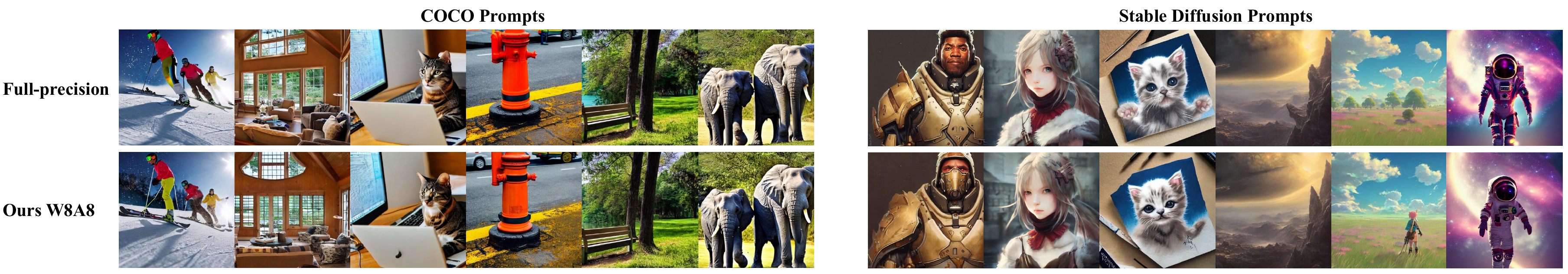}
    \caption{Stable Diffusion 512 x 512 generation results on COCO prompts and \textit{Stable-Diffusion-Prompts}. Relaxation proportion $\tau=0.20$.}
    \label{fig:sd coco vs new prompts}
\end{figure*}

An outstanding feature of text-to-image diffusion models is that they can generate images of various styles and themes, as long as the user enters the appropriate prompts. Although foundation diffusion models (e.g., Stable Diffusion and Stable Diffusion XL) are pretrained on billions of image-text pairs, researchers can usually only use a small set of data for quantization. For example, previous studies use a subset of prompts from the COCO dataset to perform calibration. In the evaluation, they only test the quantized models on the COCO prompts. However, users always input prompts with a quite different style from the calibration dataset, and generate images of various styles. Therefore, to evaluate the prompt-generalization of the quantized models,
we propose to test them also on the unseen prompts that have a quite different style from the calibration dataset. In our experiments, we evaluate the quantized model on the dataset referred to as \textit{
Stable-Diffusion-Prompts}~\cite{stable-diffusion-prompts}, which is a popular Stable Diffusion prompt dataset released on \textit{hugging face}~\cite{huggingface}. It consists of about 80,000 user-created high-quality prompts that can generate images of various styles. As shown in \cref{fig:sd coco vs new prompts}, the generated images by this dataset have quite different styles from the calibration dataset (COCO prompts in our settings), i.e.:

\textit{COCO prompts always generate realistic photographs, while Stable-Diffusion-Prompts tends to generate artistic images.}

We use the first 5,000 prompts of \textit{Stable-Diffusion-Prompts} as the prompt-generalization test set.

% \begin{figure*}
%     \centering
%     \includegraphics[width=1\linewidth]{sec/imgs/SD-hand-prompts.pdf}
%     \caption{Comparison between the quantized models and the full-precision model. Use Stable diffusion 512 x 512. Relaxation proportion $\tau=0.20$.}
%     \label{fig:sd hand prompts}
% \end{figure*}

\section{Experiments}

\subsection{Experimental Setup}

\noindent \textbf{Datasets.} 
% Since we focus on the text-to-image generation task, we need to prepare various text prompts for quantization calibration and to generate images.
Our experiments involve with two datasets, i.e., COCO~\cite{lin2014microsoft} and \textit{Stable-Diffusion-Prompts}~\cite{stable-diffusion-prompts} (as discussed in \cref{benckmark:ood}).
During the calibration, we use a few prompts of the COCO~\cite{lin2014microsoft} training dataset. During the evaluation, on the one hand, we use 5,000 prompts from the previously widely used COCO~\cite{lin2014microsoft} validation dataset. On the other hand, we further use 5,000 prompts in the \textit{Stable-Diffusion-Prompts} dataset to evaluate the prompt-generalization ability of the quantized model.

\noindent \textbf{Metrics.} We mainly adopt two metrics to evaluate generation quality. The first metric we adopt is the ``\textbf{FID to FP32}'', which is the FID score between the quantized-model generated images and full-precision-model generated images. Additionally, we also adopt the \textbf{CLIP score}~\cite{hessel2021clipscore} to evaluate whether the generated images match the text prompt. Here, we use the model ViT/L-14 to calculate CLIP scores. 
For the reference and the illustration of the problem in \cref{benckmark: fid to fp32}, we also present the metric ``FID to COCO'' adopted by previous works, but it is not proper to be used for comparison due to the distribution gap. Besides, We calculate the \textbf{BOPs}\cite{yu2020search} ($BOPs = FLOPs \cdot b_w \cdot b_a $) metric to measure computation amount, where $b_w$ and $b_a$ denote the bitwidth of weights and activations respectively. 

% \noindent \textbf{Datasets.} Our experiments involve with two datasets, i.e., COCO~\cite{lin2014microsoft} and \textit{Stable-Diffusion-Prompts} (as discussed in \cref{benckmark:ood}). COCO contains a mass of text-image pairs.
% For all experiments, we randomly sample some texts from the COCO training prompts to perform the calibration. COCO validation prompts are used to generate images for evaluation. COCO images are only used to calculate the ``FID to COCO'' score.
% \textit{Stable-Diffusion-Prompts} is a pure prompt dataset used to generate images for prompt-generalization evaluation.

% \noindent \textbf{Metrics.} We compare our proposed ``FID to FP32'' with the old ``FID to COCO'', as discussed in \cref{benckmark: fid to fp32}. We use the CLIP~\cite{hessel2021clipscore} score to measure the text-image matching.

\noindent \textbf{Baselines and implementation details.} We compare our methods with two state-of-the-art methods Q-diffusion~\cite{li2023q} and PTQ4DM~\cite{shang2023post}. For a fair comparison, we implement all the baselines and our method in a unified framework based on ~\textit{diffusers}~\cite{von-platen-etal-2022-diffusers}. For quantization, we use block reconstruction~\cite{li2021brecq} to adaptively round~\cite{nagel2020up} the weight by minimizing the mean squared errors between the quantized and the full precision outputs, following previous works~\cite{li2023q,shang2023post,he2023ptqd}. We use the Stable Diffusion v1-4  and the Stable Diffusion XL 1.0 checkpoints provided by \textit{hugging face}~\cite{huggingface}. We use the default configuration, i.e., 50-step PNDM sampling for Stable Diffusion and 50-step Euler sampling for Stable Diffusion XL. The calibration prompt number is 200 for Stable Diffusion and 30 for Stable Diffusion XL. 
% , both with the cfg scale set to 7.5. 

\noindent \textbf{Relaxation settings.} We relax the sensitive steps from 8-bit to 10-bit. By default, we set the relaxation proportion $\tau$ to 0.20 (i.e., 20\%), but we also prove relaxing only 5\% steps ($\tau=0.05$) can also outperform the baselines by a large margin.

\subsection{Main Results}
% We first compare our method with Q-diffusion and PTQ4DM on the quantization of Stable Diffusion. The generation results using the COCO prompts are present in \cref{table: coco-results}, and 
% % We randomly sample 200 prompts from the COCO training dataset to perform calibration. 
% We not only test the quantized models on the COCO validation dataset, but also test on the \textit{Stable-Diffusion-Prompts} to assess the prompt-generalization performance.

\cref{table: coco-results} shows the results of Stable Diffusion on the COCO validation prompts, which prove the effectiveness of our proposed method and benchmark.
Specifically, ``FID to COCO'' is the metric used in previous works, while ``FID to FP32'' is our proposed metric that calculates the distance between the quantized model and the full-precision 32-bit model. The quantized models have similar or even much lower ``FID to COCO'' than the full-precision model, which is inaccurate, as discussed in \cref{benckmark: fid to fp32}.
On the other hand, the results of the CLIP score and our ``FID to FP32'' show our method reaches a new state-of-the-art performance.

\begin{table}[htbp]
\caption{Quantization results on COCO validation prompts for Stable Diffusion v1-4. $\uparrow$ means higher metric is better, while $\downarrow$ means lower is better. W/A means the bitwidth for weights and activations respectively. $\tau$ denotes the proportion of activation relaxing.}
\centering
\tabcolsep=1mm
\resizebox{0.7\linewidth}{!}{
\begin{tabular}{ccccccc}
\hline
Method & Bits(W/A) & Size (GB) & BOPs (T) &  FID to COCO & FID to FP32$\downarrow$ & CLIP score$\uparrow$
\\ \hline
FP32 & 32/32  & 3.44 & 693 & 27.50 &  0.00  & 26.46 \\ \hline
Q-diffusion & 8/8 & 0.87 & 43.31 & 29.98 & 18.64 &  26.15 \\
PTQ4DM & 8/8 & 0.87 & 43.31 & 28.00 & 14.60 &  26.33 \\
PCR($\tau=0.05$) & 8/8 & 0.87 & 43.85 &  25.16 & 9.92 &  26.46 \\
PCR($\tau=0.20$) & 8/8 & 0.87 & 45.47 &  25.83 & \textbf{8.35} &  \textbf{26.47}
\\  \hline
Q-diffusion & 4/8 & 0.44 & 21.66 & 27.87  & 20.42 &  26.15 \\
PTQ4DM & 4/8 & 0.44 & 21.66 &  25.64 & 17.73 &  26.25 \\
PCR($\tau=0.05$) & 4/8 & 0.44 & 21.93 &  23.86 & 14.39 &  26.35 \\ 
PCR($\tau=0.20$) & 4/8 & 0.44 & 22.74 &  22.04 & \textbf{14.25} &  \textbf{26.48}
\\  \hline
% PCR(Ours) & 8/32 & 43.31 &  27.58 & 2.95 &  26.47 \\
% PCR(Ours) & 4/32 & 43.31 & 24.29 & 12.45 &  26.50
% \\  \hline
\end{tabular}
}
\label{table: coco-results}
\end{table}

\begin{table}[htbp]
\caption{Quantization results on COCO validation prompts for Stable Diffusion XL.}
\tabcolsep=1mm
\centering
\resizebox{0.7\linewidth}{!}{
\begin{tabular}{ccccccc}
\hline
Method & Bits(W/A) & Size (GB) & BOPs (T) &  FID to COCO & FID to FP32$\downarrow$ & CLIP score$\uparrow$
\\ \hline
FP32 & 32/32 & 10.30 & 6933 & 27.28 &  0.00  & 26.44 \\ \hline
Q-diffusion & 8/8 & 2.61 & 433 &  42.85 & 38.19 &  15.97 \\
PTQ4DM & 8/8 & 2.61 & 433 & 42.33 & 38.65 &  15.98 \\
PCR($\tau=0.05$) & 8/8 & 2.61 & 438 &  30.87 & 22.08 &  19.73 \\
PCR($\tau=0.20$) & 8/8 & 2.61 & 455 &  26.16 & \textbf{12.00} &  \textbf{24.05}
\\  \hline
Q-diffusion & 4/8  & 1.32 & 217 &  45.43 & 44.07 & 15.91 \\
PTQ4DM & 4/8 & 1.32 & 217 &  43.37 & 46.45 &  15.92 \\
PCR($\tau=0.05$) & 4/8 & 1.32 & 219 & 34.99 & 30.87 &  19.51 \\
PCR($\tau=0.20$) & 4/8 & 1.32 & 227 & 24.27 & \textbf{18.27} &  \textbf{23.85}
\\  \hline
\end{tabular}
}
\label{table: coco-results-sdxl}
\end{table}

\textbf{The discussion about CLIP score.} \cref{table: coco-results} also reveals that the CLIP score might not be able to precisely distinguish the quality of quantized models when their CLIP score is closer to the full-precision score. For example, the CLIP score of PCR 4/8 is a little higher than PCR 8/8 ($\tau=0.20$), even though the quality of PCR 8/8 ($\tau=0.20$), is obviously better, as shown in \cref{fig:appendix sd on coco} and \cref{fig:appendix sd on new prompts} in the Appendix.
% \cref{table: ablation-1} also indicates the similar circumstance. 
These phenomena support the significance of our ``FID to FP32'' metric to measure image fidelity. Therefore, it is significant to combine the CLIP score and our ``FID to FP32'' score to comprehensively measure the quality of quantized models.
% TODO  可视化图，放到附录

% To further validate our method for the larger model and higher resolution, we experiment on Stable Diffusion XL, generating images with a resolution of 768 $\times$ 768.

\cref{table: coco-results-sdxl} show the results of Stable Diffusion XL on COCO validation prompts. Our method outperforms previous methods by an obvious margin. 
Moreover, the visualized results in \cref{fig:sdxl comparison} show that previous methods fail to generate images that can properly match the prompt, while our method can generate high-quality images matching the text.

\cref{table: coco-results} and \cref{table: coco-results-sdxl} also indicate relaxing only 5\% steps ($\tau=0.05$) can also outperform the baselines by a large margin.

To assess prompt generalization, we directly test quantized checkpoints (obtained on COCO) on the unseen dataset \textit{Stable-Diffusion-Prompts}. \cref{table: sd-ood-prompts} show the results of Stable Diffusion. Although our method has a slightly higher FID score in the setting of W4/A8, its CLIP score obviously exceeds the baselines. \cref{table: sdxl-ood-prompts} show the results of Stable Diffusion XL. 
These results prove our method has better generalization on the unseen prompts outside the calibration dataset.

\begin{table}[htbp]
\caption{Results on \textit{Stable-Diffusion-Prompts} dataset for Stable Diffusion v1-4.}
\tabcolsep=1mm
\centering
\resizebox{0.57\linewidth}{!}{
\begin{tabular}{cccccc}
\hline
Method & Bits(W/A) & Size (GB) & BOPs (T) & FID to FP32$\downarrow$ & CLIP score$\uparrow$
\\ \hline
FP32 & 32/32 & 3.44 & 693 &  0.00  & 28.79 \\ \hline
Q-diffusion & 8/8  & 0.87 & 43.31 & 16.22 &  27.19 \\
PTQ4DM & 8/8  & 0.87 & 43.31 & 13.25 &  27.77 \\
PCR($\tau=0.20$) & 8/8  & 0.87  & 45.47 & \textbf{9.52} &  \textbf{28.74}
\\  \hline
Q-diffusion & 4/8 & 0.44 & 21.66 & 17.43 &  27.31 \\
PTQ4DM & 4/8  & 0.44 & 21.66 & \textbf{17.28} &  27.39 \\
PCR($\tau=0.20$) & 4/8  & 0.44 & 22.74 & 17.95 &  \textbf{28.05}
\\  \hline
\end{tabular}
}
\label{table: sd-ood-prompts}
\end{table}

\begin{table}[htbp]
\caption{Results on \textit{Stable-Diffusion-Prompts} dataset for Stable Diffusion XL.}
\tabcolsep=1mm
\centering
\resizebox{0.57\linewidth}{!}{
\begin{tabular}{cccccc}
\hline
Method & Bits(W/A) & Size (GB)  & BOPs (T) & FID to FP32$\downarrow$ & CLIP score$\uparrow$
\\ \hline
FP32 & 32/32 & 10.30 & 6933 &  0.00  & 29.74 \\ \hline
Q-diffusion & 8/8 & 2.61 & 433 & 22.81 &  20.01 \\
PTQ4DM & 8/8 & 2.61   & 433 & 22.85 &  20.09 \\
PCR($\tau=0.20$) & 8/8 & 2.61  & 455 & \textbf{11.71}  &  \textbf{25.74}
\\  \hline
Q-diffusion & 4/8  & 1.32 & 217  & 24.98 &  19.93 \\
PTQ4DM & 4/8  & 1.32 & 217  & 28.28 &  20.13 \\
PCR($\tau=0.20$) & 4/8 & 1.32  & 227  & \textbf{18.25} &  \textbf{25.71}
\\  \hline
\end{tabular}
}
\label{table: sdxl-ood-prompts}
\end{table}

Comparing the results evaluated on COCO with the results on \textit{Stable-Diffusion-Prompts}, we can observe some differences. For example, the gap between these quantized Stable Diffusion XL models is obviously smaller on the ~\textit{Stable-Diffusion-Prompts} than the COCO. This may indicate slight overfitting to the calibration data, which also supports the necessity of the evaluation on the unseen prompts outside the calibration data.

\begin{figure}
    \centering
    \includegraphics[width=0.8\linewidth]{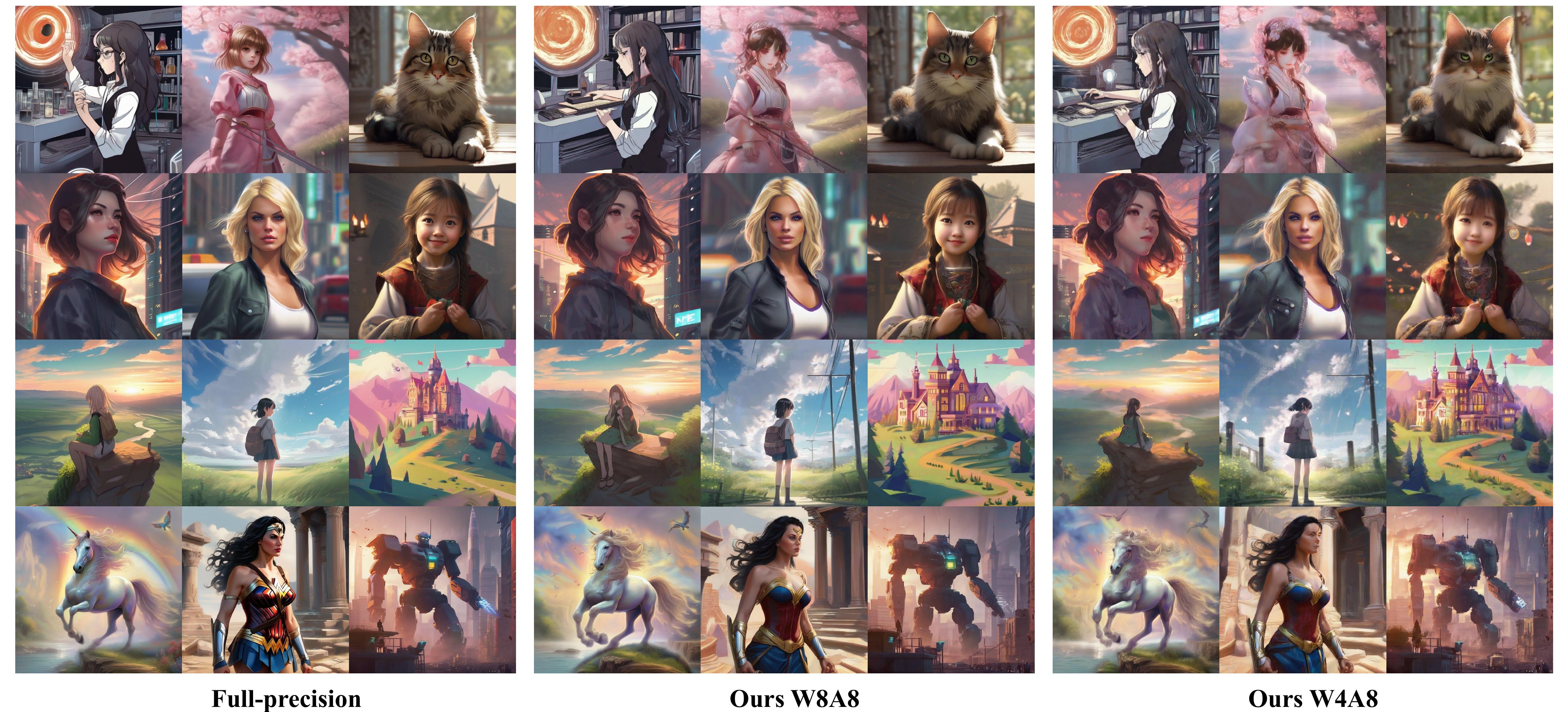}
    \caption{Stable Diffusion XL 768x768 generation using \textit{Stable-Diffusion-Prompts}.}
    \label{fig:appendix sdXL on new prompts}
\end{figure}

\cref{fig:appendix sdXL on new prompts} shows the generated samples by Stable Diffusion XL using \textit{Stable-Diffusion-Prompts}. More samples can be found in the Appendix.

The BOPs results in~\cref{table: coco-results,table: coco-results-sdxl,table: sd-ood-prompts,table: sdxl-ood-prompts} indicate that our PCR method only has a negligible extra computational cost. %TODO fairness

\begin{table}[htbp]
\caption{Ablation results on the COCO validation prompts for Stable Diffusion v1-4. ``Relax'' denotes the activation relaxing, and ``Prog'' denotes the progressive calibration.}
\tabcolsep=1mm
\centering
\resizebox{0.55\linewidth}{!}{
\begin{tabular}{cccc}
\hline
Method & Bits(W/A)  & FID to FP32$\downarrow$ & CLIP score$\uparrow$ 
\\ \hline
FP32 & 32/32  &  0.00  & 26.46 \\ \hline
Base  & 8/8  & 15.90 &  26.42  \\
+ Relax & 8/8  & 9.02 &  26.42  \\
+ Prog & 8/8  & 11.01 & \textbf{26.48}  \\
+ Prog \& Relax & 8/8 & \textbf{8.35} &  26.47 
\\  \hline
Base & 4/8  & 25.29 &  26.21  \\
+ Relax & 4/8  & 15.63 &  26.35  \\
+ Prog & 4/8  & 17.54 & \textbf{26.49} \\
+ Prog \& Relax & 4/8  &  \textbf{14.25} &  26.48 
\\  \hline
\end{tabular}
}
\label{table: ablation-1}
\end{table}

\subsection{Ablation Study}
\label{ablation study}

\textbf{The effects of our components.}
In this section, we analyze the effects of different components. \cref{table: ablation-1} demonstrates that solely using the activation relaxing and solely using the progressive calibration can both improve the performance a lot. And combining these two strategies further improves the quality. Moreover, note that the progressive calibration is essential for maintaining the CLIP score. The quantized models suffer a remarkable degradation of CLIP score without the progressive calibration. For example, in the W4A8 setting, only adding the activation relaxing to the base just has a 26.35 CLIP score (row 8), while further adding the progressive calibration can raise the score to 26.48 (row 10).

\begin{figure}[htbp]
    \centering
    \includegraphics[width=0.55\linewidth]{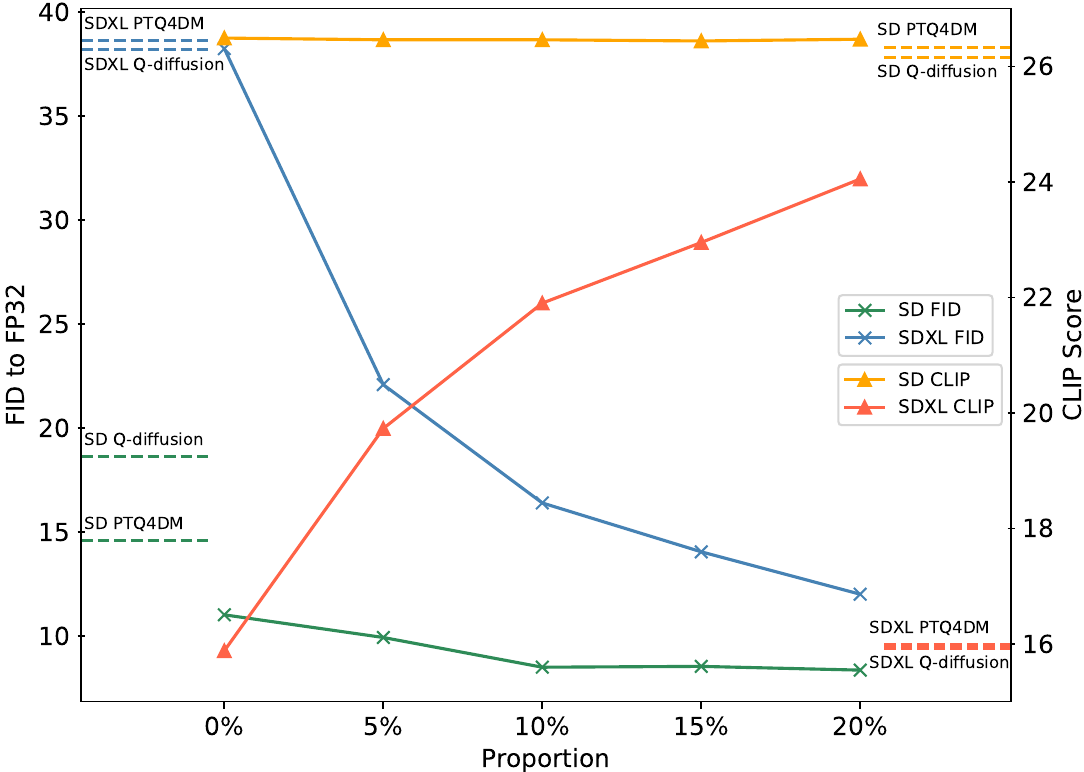}
    \caption{Score variations \textit{w.r.t.} the relaxing proportion, on COCO validation prompts, W8/A8. The solid lines denote the scores of PCR. The dashed lines parallel to the x-axis denote the corresponding scores of the baselines.}
    \label{fig: proportion}
\end{figure}

\noindent \textbf{The effects of the relaxing proportion.}
We explore how the relaxation proportion $\tau$ affects the performance. \cref{fig: proportion} shows the score changes of Stable Diffusion (SD) and Stable Diffusion XL (SDXL) with respect to the proportion.
The results show the gain from relaxing more steps decreases rapidly as the proportion grows, which means the steps nearest the $x_0 / x_T$ are the most important to the image-fidelity/text-image-matching.
% It also indicates relaxing only 5\% steps can also obviously outperform the baselines.
%It indicates the steps near the end 

\section{Conclusion}

This paper aims to advance the quantization of text-to-image diffusion models from two perspectives, method and benchmark. We propose a novel quantization method PCR and an effective benchmark QDiffBench. Extensive experiments prove the superiority of our method and benchmark.

\section*{Acknowledgement}

This work is supported by the National Key Research and Development Program of China No.2023YFF1205001, National Natural Science Foundation of China (No. 62250008, 62222209,  62102222), Beijing National Research Center for Information Science and Technology under Grant No. BNR2023RC01003, BNR2023TD03006, and Beijing Key Lab of Networked Multimedia.

% \clearpage  % TODO REVIEW/FINAL: This \clearpage needs to be removed from both review and camera-ready versions.

% ---- Bibliography ----
%
% BibTeX users should specify bibliography style 'splncs04'.
% References will then be sorted and formatted in the correct style.
%
\bibliographystyle{splncs04}
\bibliography{main}

%############################################################# appendix #####################################################

\renewcommand{\thetable}{A.\arabic{table}}
\renewcommand{\thefigure}{A.\arabic{figure}}
\renewcommand{\thesection}{A.\arabic{section}}

\clearpage
\setcounter{page}{1}

\title{Supplementary Material} 

\author{}
\institute{}

\maketitle

\section{Proof of Theorem 1}
\label{appdedix: proof}
The section will prove \cref{theorem:1} in~\cref{sec: progressive calibration}. We first review the notations and the theorem, and then give the proof.

\noindent \textbf{Notations.} Let $\boldsymbol{\epsilon}_{\theta}$ be the denoising net and $\mathbf{x}_t$ be the intermediate variable at timestep $t$. Let the mark $\hat{..}$ denote the quantized term, and $\hat{\boldsymbol{\epsilon}}_{\theta}\left(\hat{\mathbf{x}}_{t}, t\right)=\boldsymbol{\epsilon}_{\theta}\left(\hat{\mathbf{x}}_{t}, t\right)+\Delta_{t}$, where $\boldsymbol{\epsilon}_{\theta}\left(\hat{\mathbf{x}}_{t}, t\right)$ is the predicted noise with quantized input but full-precision network, and $\hat{\boldsymbol{\epsilon}}_{\theta}\left(\hat{\mathbf{x}}_{t},t\right)$ is the predicted noise with both quantized input and network. Without loss of generality, we use the DDPM sampler~\cite{ho2020denoising} there, i.e.,

\begin{align}
\mathbf{x}_{t-1}=\frac{1}{\sqrt{\alpha_{t}}}\left(\mathbf{x}_{t}-\frac{1-\alpha_{t}}{\sqrt{1-\bar{\alpha}_{t}}} \boldsymbol{\epsilon}_{\theta}\left(\mathbf{x}_{t}, t\right)\right)+\sigma_{t} \mathbf{z} , \nonumber
\end{align}

\noindent where $\alpha_{t}$ and $\bar{{\alpha}_{t}}$ are constants of the forward process, and $z$ is a random gaussian variable.

\begin{theorem}
Through the multi-step denoising process, the upper bound of the ultimate quantization error $\delta = \|\mathbf{x}_0 - \mathbf{\hat{x}}_0\|$ can be approximated as the linear combination of $\{\|\Delta_t\|\}, \ t=1,..,T.$
\end{theorem}

% \begin{proof}
% \end{proof}

\noindent \textit{Proof.} Without loss of generality, we set $\sigma_{t}$ to 0. Consider the first step of the sampling process, i.e., from $\mathbf{x}_T$ to $\mathbf{x}_{T-1}$:

{ \footnotesize
\begin{align}
\mathbf{x}_{T-1}&=\frac{1}{\sqrt{\alpha_T}}\left(\mathbf{x}_T-\frac{1-\alpha_T}{\sqrt{1-\bar{\alpha}_T}}\boldsymbol{\epsilon}_\theta(\mathbf{x}_T,T)\right).
\nonumber
\end{align}
}

\noindent After quantization, it derives that:

{  \footnotesize
\begin{align}
\label{eq:1}
\hat{\mathbf{x}}_{T-1} &=\frac1{\sqrt{\alpha_T}}\left(\hat{\mathbf{x}}_T-\frac{1-\alpha_T}{\sqrt{1-\bar{\alpha}_T}}\left(\boldsymbol{\epsilon}_\theta\left(\hat{\mathbf{x}}_T,T\right)+\Delta_T\right)\right)
\nonumber
\\ & \overset{(a)}{=} \frac1{\sqrt{\alpha_T}}\left(\mathbf{x}_T-\frac{1-\alpha_T}{\sqrt{1-\bar{\alpha}_T}}\left(\boldsymbol{\epsilon}_\theta\left(\mathbf{x}_T,T\right)+\Delta_T\right)\right)
\nonumber
\\ &= \mathbf{x}_{T-1} - \frac1{\sqrt{\alpha_T}}\frac{1-\alpha_T}{\sqrt{1-\bar{\alpha}_T}} \Delta_T  .
% \nonumber
\end{align}
}

\noindent The equation (a) holds because $\hat{\mathbf{x}}_T = \mathbf{x}_T$ which is sampled from $\mathcal{N}(\mathbf{0}, \mathbf{I})$.
Similarly, we next consider the step from $\mathbf{x}_{T-1}$ to $\mathbf{x}_{T-2}$:

{ 
% \scriptsize
\footnotesize
\begin{align}
\label{eq:2}
\hat{\mathbf{x}}_{T-2} &=\frac1{\sqrt{\alpha_{T-1}}}\Bigg(\hat{\mathbf{x}}_{T-1}-\frac{1-\alpha_{T-1}}{\sqrt{1-\bar{\alpha}_{T-1}}}\Big(\boldsymbol{\epsilon}_\theta\left(\hat{\mathbf{x}}_{T-1},{T-1}\right)+\Delta_{T-1}\Big)\Bigg)
\nonumber
\\ & \overset{(b)}{\approx} \frac1{\sqrt{\alpha_{T-1}}}\Bigg(\hat{\mathbf{x}}_{T-1}-\frac{1-\alpha_{T-1}}{\sqrt{1-\bar{\alpha}_{T-1}}}\Big(\boldsymbol{\epsilon}_\theta\left({\mathbf{x}}_{T-1},{T-1}\right) \nonumber
\\  &  \qquad \qquad \qquad  -\boldsymbol{\epsilon}_\theta^{\prime}(\mathbf{x}_{T-1},T-1)  \frac1{\sqrt{\alpha_T}}\frac{1-\alpha_T}{\sqrt{1-\bar{\alpha}_T}} \Delta_T +\Delta_{T-1}\Big)\Bigg)
\end{align}
}

\noindent where (b) is derived from Taylor expansion, i.e.,

{ \footnotesize
\begin{align}
\boldsymbol{\epsilon}_\theta\left(\hat{\mathbf{x}}_{T-1},{T-1}\right) &\approx \boldsymbol{\epsilon}_\theta\left({\mathbf{x}}_{T-1},{T-1}\right) \nonumber
\\ & \quad -\boldsymbol{\epsilon}_\theta^{\prime}(\mathbf{x}_{T-1},T-1) \frac1{\sqrt{\alpha_T}}\frac{1-\alpha_T}{\sqrt{1-\bar{\alpha}_T}} \Delta_T  \nonumber
% \\ & \quad - o( \frac1{\sqrt{\alpha_T}}\frac{1-\alpha_T}{\sqrt{1-\bar{\alpha}_T}} \Delta_T) 
. \nonumber
\end{align}
}

\noindent Replacing $\hat{\mathbf{x}}_{T-1}$ with \cref{eq:1}, \cref{eq:2} can be written as:

{ \footnotesize
\begin{align}
\label{eq:3}
\hat{\mathbf{x}}_{T-2} & \overset{(b)}{\approx} \frac1{\sqrt{\alpha_{T-1}}}\Bigg( \mathbf{x}_{T-1} - \frac1{\sqrt{\alpha_T}}\frac{1-\alpha_T}{\sqrt{1-\bar{\alpha}_T}} \Delta_T  \nonumber
\\ & \qquad \qquad \qquad -\frac{1-\alpha_{T-1}}{\sqrt{1-\bar{\alpha}_{T-1}}}\Big(\boldsymbol{\epsilon}_\theta\left(\mathbf{x}_{T-1},{T-1}\right) \nonumber
\\  &  \qquad \qquad \qquad  -\boldsymbol{\epsilon}_\theta^{\prime}(\mathbf{x}_{T-1},T-1)  \frac1{\sqrt{\alpha_T}}\frac{1-\alpha_t}{\sqrt{1-\bar{\alpha}_t}} \Delta_T +\Delta_{T-1}\Big)\Bigg)  \nonumber
\\ &=  \mathbf{x}_{T-2} - \frac1{\sqrt{\alpha_T}} \frac1{\sqrt{\alpha_{T-1}}} \cdot \frac{1-\alpha_T}{\sqrt{1-\bar{\alpha}_T}} \Delta_T  \nonumber \\
& \quad + \frac1{\sqrt{\alpha_T}}\frac1{\sqrt{\alpha_{T-1}}} \cdot \frac{1-\alpha_T}{\sqrt{1-\bar{\alpha}_T}} \frac{1-\alpha_{T-1}}{\sqrt{1-\bar{\alpha}_{T-1}}} \boldsymbol{\epsilon}_\theta^{\prime}(\mathbf{x}_{T-1},T-1) \Delta_T  \nonumber \\
& \quad -\frac1{\sqrt{\alpha_{T-1}}}\frac{1-\alpha_{T-1}}{\sqrt{1-\bar{\alpha}_{T-1}}} \Delta_{T-1}  .
\end{align}
}

\noindent \cref{eq:1} indicates $\mathbf{x}_{T-1} - \hat{\mathbf{x}}_{T-1}$
is the linear combination of $\Delta_T$. Similarly, \cref{eq:3} proves $\mathbf{x}_{T-2} - \hat{\mathbf{x}}_{T-2}$
is the linear combination of $\Delta_T$ and $\Delta_{T-1}$. Iteratively, we can prove $ \mathbf{x}_0 - \hat{\mathbf{x}}_0$ is the linear combination of $\{\Delta_t\}, \ t=1,..,T$. Then, we can easily derive that the upper bound of $\delta = \|\mathbf{x}_0 - \hat{\mathbf{x}}_0\|$ is the linear combination of $\{\|\Delta_t\|\}, \ t=1,..,T.$

\section{Additional Visual Results}

\begin{figure*}[!htbp]
    \centering
    \includegraphics[width=1\linewidth]{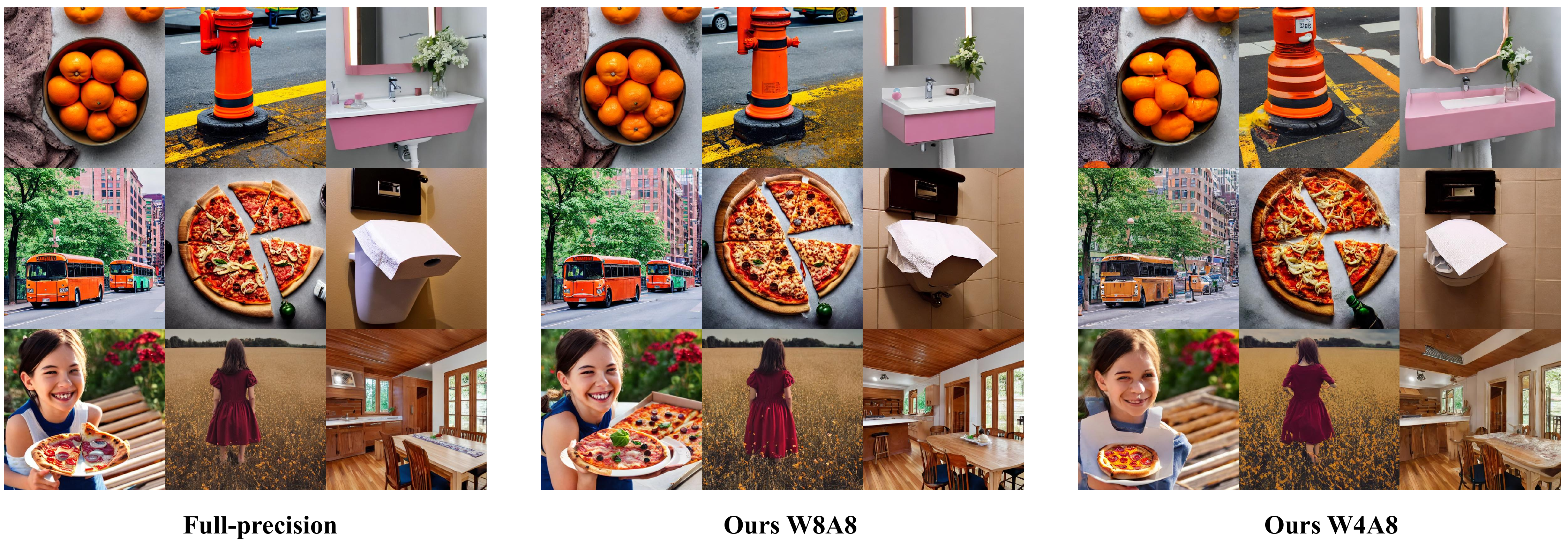}
    \caption{Stable Diffusion image generation using COCO prompts.The relaxation proportion $\tau=0.20$.}
    \label{fig:appendix sd on coco}
\end{figure*}

\begin{figure*}[!htbp]
    \centering
    \includegraphics[width=1\linewidth]{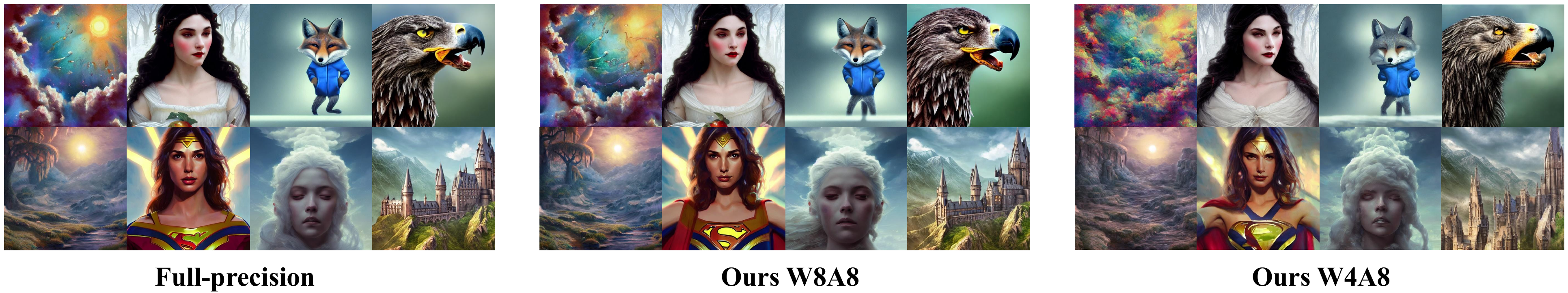}
    \caption{Stable Diffusion image generation using \textit{Stable-Diffusion-prompts}. The relaxation proportion $\tau=0.20$.}
    \label{fig:appendix sd on new prompts}
\end{figure*}

\begin{figure*}[!htbp]
    \centering
    \includegraphics[width=1\linewidth]{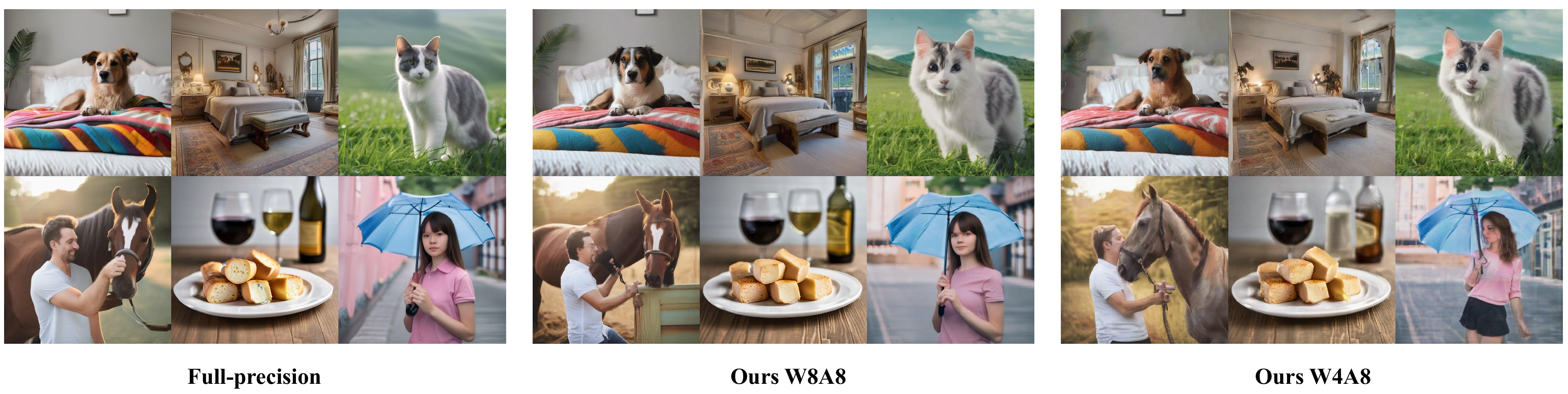}
    \caption{Stable Diffusion XL image generation using COCO prompts. The relaxation proportion $\tau=0.20$.}
    \label{fig:appendix sdXL on coco}
\end{figure*}

We show more visualized results of quantized models on COCO prompts and \textit{Stable-Diffusion-prompts}, i.e., \cref{fig:appendix sd on coco,fig:appendix sd on new prompts,fig:appendix sdXL on coco}. The COCO prompts tend to generate realistic photographs, while ~\textit{Stable-Diffusion-Prompts} tends to generate artistic images. 

% We also present the pseudo-code of our proposed PCR method in \cref{algorithm: pcr}.

\section{Additional Results of Unconditional Diffusion}

We also validate our method in the unconditional scenarios, even though it is not our major concern. 
We use the pretrained checkpoint \textit{CompVis/ldm-celebahq-256} released on \textit{hugging face}. We generate 50,000 images and calculate the FID to the original training data, i.e., CelebA-HQ. The results are shown in \cref{table: ldm-celehq }, which indicates our method is also effective for unconditional diffusion models.

\begin{table}[htbp]
\caption{Results on the CelebA-HQ using the unconditional latent diffusion model. Use DDIM 100-step sampler.}
\tabcolsep=1mm
\centering
\resizebox{0.32\linewidth}{!}{
\begin{tabular}{ccc}
\hline
Method & Bits(W/A)  & FID $\downarrow$
\\ \hline
FP32 & 32/32  &  12.24  \\ \hline
Q-diffusion & 8/8  & 18.43  \\
PTQ4DM & 8/8  & 17.18  \\
PCR(Ours) & 8/8 & \textbf{13.22}
\\  \hline
Q-diffusion & 4/8  & 22.59  \\
PTQ4DM & 4/8  & 22.15  \\
PCR(Ours) & 4/8  & \textbf{19.38} 
\\  \hline
\end{tabular}
}
\label{table: ldm-celehq }
\end{table}

\section{Additional Results on DiffusionDB datasets}

To further evaluate the prompt-generalization ability of the model, we also conducted experiments on the DiffusionDB dataset. Like \textit{Stable-Diffusion-Prompts}, DiffusionDB contains human-created prompts, which simulate the situation in practical uses. The results are shown in \cref{table: sd-diffusiondb-prompts}, which indicates our method performs excellently on this prompt-generalization dataset.

\begin{table}[htbp]
\caption{Results on the \textit{DiffusionDB} dataset for Stable Diffusion 512 x 512.}
\tabcolsep=1mm
\centering
\resizebox{0.5\linewidth}{!}{
\begin{tabular}{cccc}
\hline
Method & Bits(W/A)  & FID to FP32 $\downarrow$ & CLIP score $\uparrow$
\\ \hline
FP32 & 32/32  &  0.00  & 29.21\\ \hline
Q-diffusion & 8/8  & 16.30&  27.98\\
PTQ4DM & 8/8  & 14.26&  28.41\\
PCR(Ours) & 8/8 & \textbf{10.03}&  \textbf{29.20}\\  \hline
Q-diffusion & 4/8  & 18.77&  27.84\\
PTQ4DM & 4/8  & \textbf{17.84}&  28.07\\
PCR(Ours) & 4/8  & 18.10&  \textbf{28.65}\\  \hline
\end{tabular}
}
\label{table: sd-diffusiondb-prompts}
\end{table}

\section{Further Analysis about Distribution Gap}
\label{sec: analysis about distribution gap}

\begin{figure*}[!htbp]
    \centering
    \includegraphics[width=0.8\linewidth]{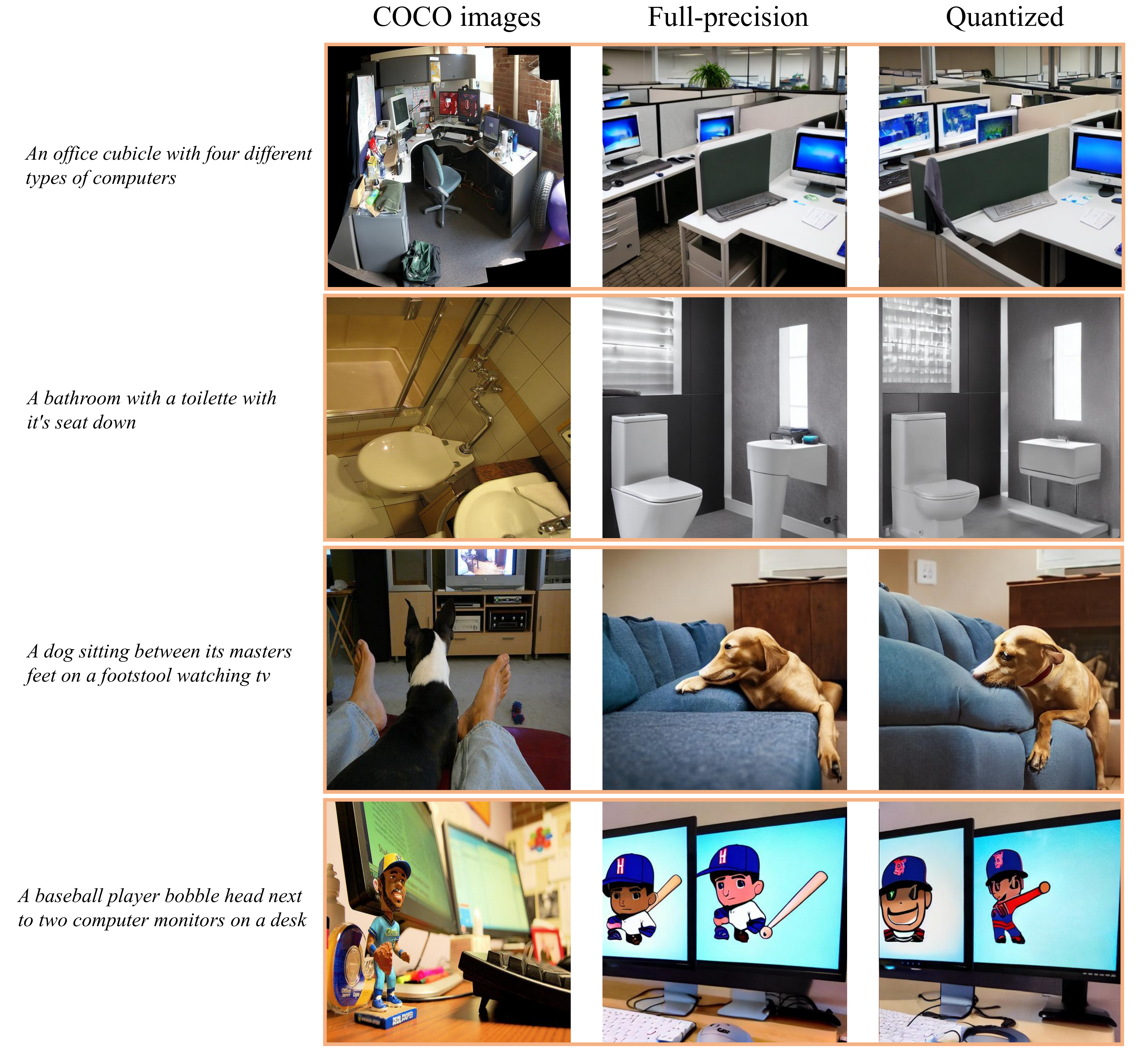}
    \caption{The comparison between the COCO images and the Stable Diffusion generated images.  Row 1 and 2 show that COCO images are photorealism, while Stable Diffusion can't achieve this well. Row 3 and 4 show that Stable Diffusion can sometimes be challenging to generate images with perfect text-image matching.}
    \label{fig:appendix coco vs. generated}
\end{figure*}

We have pointed out the reason why ``FID to COCO'' is inaccurate is the data distribution-gap between COCO images and diffusion-model-generated images. We believe that the distribution-gap may result from two aspects: i) Large text-to-image diffusion models are pretrained on the large-scale dataset, e.g., LAION-5B, whose distribution quite differs from COCO. ii) Additionally, limited by the text-to-image generative ability of the model, generating high-quality images that perfectly match the prompt can sometimes be challenging, while text-image matching of COCO is good because of the human annotations. Moreover, there are fake details more or less in the generated images, especially for human faces, keyboards, and so on.
We may need to try different random seeds many times to produce a satisfying image. We demonstrate this statement with \cref{fig:appendix coco vs. generated}. Row 1 and 2 show the distribution gap in the form of style difference. Row 3 and 4 show that the generated images are not satisfactory because of not matching the prompt well.

\section{More FID-to-FP32 Results Based on Recent Feature}

Kynkaanniemi et al.~\cite{kynkaanniemi2022role} indicate the original Frechet distance based on the Inception network might have some biases. Therefore, we supplement the FID-to-FP32 scores based on CLIP and DINO-v2 features. The results of SD-1.4 with 50-step PNDM are shown in \cref{table: clip-based fid}. ~\cref{table: lcm,table: unipc} also report these metrics. These results show the recent metrics are fully consistent with the original FID-to-FP32 score in the context of diffusion model quantization. We don't observe any bias in this setting.

\begin{table}[htbp]
\caption{Results of recent-feature based FID-to-FP32 on COCO.}
\centering
\tabcolsep=1mm
\resizebox{\linewidth}{!}{
\begin{tabular}{cccccc}
\toprule
Method & Bits  (W/A) & FID to FP32  (ori.) $\downarrow$ &  FID to FP32   (clip) $\downarrow$ &  FID to FP32  (dino) $\downarrow$ & CLIP$\uparrow$
\\ \midrule
FP32 & 32/32  & 0.00  & 0.00 &  0.00 & 26.46 \\ \midrule
Q-diffusion & 8/8  & 18.64 & 13.61 & 152.6 &  26.15 \\
PTQ4DM & 8/8  & 14.60 & 9.87 & 123.0 &  26.33 \\
Ours($\tau=0.20$) & 8/8 & \textbf{8.35} & \textbf{3.04} & \textbf{67.6}  &  \textbf{26.47}
\\ 
% \midrule
% Q-diffusion & 4/8  & 20.42 & 18.74 & 174.3 &  26.15 \\
% PTQ4DM & 4/8 & 17.73 & 16.19 &  152.1 &  26.25 \\
% Ours($\tau=0.20$) & 4/8 & \textbf{14.25} & \textbf{9.55} & \textbf{125.8} &  \textbf{26.48}
% \\
\bottomrule
\end{tabular}
}
\label{table: clip-based fid}
\end{table}

\section{Effectiveness with very few sampling steps.}

We test our method with advanced few-step samplers, i.e., Euler-30 and UniPC-10, whose results are shown in \cref{table: unipc}. Besides, we test the LCM~\cite{luo2023latent} model with 5 steps, as shown in \cref{table: lcm}. These results show our method works well for various and few-step sampling.

\begin{table}[htbp]
\caption{Results of SD with Euler and UniPC on COCO, W8A8.}
\centering
\tabcolsep=1mm
\resizebox{1\linewidth}{!}{
\begin{tabular}{cccccc}
\toprule
Sampler & method & FID to FP32 (ori.) $\downarrow$ &  FID to FP32 (clip) $\downarrow$ &  FID to FP32 (dino) $\downarrow$ & CLIP score$\uparrow$
\\ \midrule
\multirow{3}*{Euler-30}  & FP32  & 0.00  & 0.00 &  0.00 & 26.62 \\ 
% & Q-diffusion   &  &   &  &  \\
& PTQ4DM   & 10.16  &  4.95 & 87.6 & 26.47 \\ 
&Ours($\tau=0.20$)  &  \textbf{8.34} & \textbf{2.98}  & \textbf{67.2} &  \textbf{26.57}

\\ 
\midrule
\multirow{3}*{UniPC-10}  & FP32  & 0.00  & 0.00 &  0.00 & 26.46 \\ 
% & Q-diffusion   &  & &  &   \\
& PTQ4DM   & 30.95 & 23.03 &   253.7 &  25.03  \\
&Ours($\tau=0.20$)  & \textbf{8.98} & \textbf{3.30} & \textbf{75.6} &  \textbf{26.41}
\\ 
\bottomrule
\end{tabular}
}
\label{table: unipc}
\end{table}

\begin{table}[htbp]
\caption{Results of LCM with 5 steps on COCO dataset.}
\tabcolsep=1mm
\centering
\resizebox{1\linewidth}{!}{
\begin{tabular}{cccccc}
\toprule
Method & Bits(W/A)  & FID to FP32 (ori.) $\downarrow$ &  FID to FP32 (clip) $\downarrow$ &  FID to FP32 (dino) $\downarrow$ & CLIP score$\uparrow$ 
\\ \hline
FP32 (LCM) & 32/32  &  0.00 & 0.00 & 0.00 & 25.16 \\ \midrule
PTQ4DM  & 8/8  & 10.13 & 5.21 &  98.33 &  25.06 \\
% Ours w/o Prog & 8/8  &  & & & \textbf{26.48}  \\
Ours($\tau=0.20$)  & 8/8 & \textbf{8.40} & \textbf{4.14} &  \textbf{82.7} &  \textbf{25.23}
\\  \midrule
PTQ4DM & 4/8  & 16.80 & 16.03 &  178.5 & 24.73  \\
% Ours w/o Prog & 4/8  &  & & & \textbf{26.49} \\
Ours($\tau=0.20$)  & 4/8  & \textbf{14.63} & \textbf{12.88}  &  \textbf{147.9} & \textbf{24.87}
\\  \bottomrule
\end{tabular}
}
\label{table: lcm}
\end{table}

\renewcommand{\algorithmicrequire}{\textbf{Input:}}
\renewcommand{\algorithmicensure}{\textbf{Output:}}
\begin{algorithm}
\caption{Our proposed PCR algorithm}
\label{algorithm: pcr}
\begin{algorithmic}[1]
\REQUIRE The full-precision model $W_\theta$, Sampling steps $T$
\REQUIRE Relaxing timesteps list $R$, activation bitwidths $B$
\REQUIRE The calibration prompts $\mathcal{P}$ selected from COCO
\ENSURE Quantized weight $\hat{W}_\theta$ and activation quantizers $\{q_i\}, i=1,...,T$
\STATE Calibrate the weight quantizers with $W_\theta$ and $\mathcal{P}$. Update quantized weight $\hat{W}_\theta$ 
\STATE \textbf{Activation relaxing:}
\FOR{$i=1$ to $T$}
    \IF{$i$ in $R$}
        \STATE set $B[i]$ to higher bit
    \ENDIF
\ENDFOR

\STATE \textbf{Progressive Calibration:}

\FOR{$t=T$ to $1$}
    \STATE generate calibration data at $t$  with $\hat{W}_\theta$, $\mathcal{P}$, and updated quantizer $\{q_i\}, i=t+1,...,T$ of previous steps
    \STATE calibrate and update the quantizer at t, i.e., $q_t$, with bitwidth $B[t]$
\ENDFOR
\end{algorithmic}
\end{algorithm}

\end{document}